\documentclass[11pt]{article}

\usepackage[]{EMNLP2023}

\usepackage{times}
\usepackage{latexsym}
\usepackage{graphicx}
\usepackage{rotating}
\usepackage[T1]{fontenc}
\usepackage[utf8]{inputenc}
\usepackage{xspace}

\usepackage{adjustbox} 
\usepackage{supertabular}

\usepackage[T1]{fontenc}

\usepackage[utf8]{inputenc}

\usepackage{microtype}

\usepackage{inconsolata}

\newcommand{\camerareadytext}[1]{\xspace}
\newcommand{\Sref}[1]{\S\ref{#1}}

\newcommand{\biberplus}{\textsc{BiberPlus}\xspace}
\newcommand{\neurobiber}{\textsc{NeuroBiber}\xspace}

\title{Neurobiber: Fast and Interpretable Stylistic Feature Extraction}

\author{
  Kenan Alkiek \\
  University of Michigan \\ 
  \texttt{kalkiek@umich.edu} \\
  \And
  Anna Wegmann \\
  Utrecht University \\ 
  \texttt{a.m.wegmann@uu.nl}
  \AND
  Jian Zhu \\
  University of British Columbia \\ 
  \texttt{jian.zhu@ubc.ca} \\
  \And
  David Jurgens \\
  University of Michigan \\ 
  \texttt{jurgens@umich.edu}
}

\begin{document}
\maketitle
\begin{abstract}
Linguistic style is pivotal for understanding how texts convey meaning and fulfill communicative purposes, yet extracting detailed stylistic features at scale remains challenging. We present \neurobiber, a transformer-based system for fast, interpretable style profiling built on Biber’s Multidimensional Analysis (MDA). \neurobiber\ predicts 96 Biber-style features from our open-source \biberplus\ library—a Python toolkit that computes stylistic features and provides integrated analytics (e.g., PCA, factor analysis). Despite being up to 56 times faster than existing open source systems, \neurobiber\ replicates classic MDA insights on the CORE corpus and achieves competitive performance on the PAN 2020 authorship verification task without extensive retraining. Its efficient and interpretable representations readily integrate into downstream NLP pipelines, facilitating large-scale stylometric research, forensic analysis, and real-time text monitoring. All components are made publicly available.

\end{abstract}

\section{Introduction}
\label{sec:intro}

The style in which text is written plays an instrumental role in how a message can be understood \cite{coupland2007style,eckert2012three}. %
This style, in part, situates the text within a broader context and social setting based on recognized patterns in style, e.g., the use of a formal writing style, the register of conversation or news, or styles used widely by certain ages. Quantifying style computationally requires some method for \textit{representing style} and comparing these representations for different texts. Such style representations can benefit a broad spectrum of tasks, such as style transfer \cite{shen_style_2017,fu_style_2017,dai_style_2019,li_domain_2019,yi_text_2020,zhu_styleflow_2022}, and machine translation \cite{niu_multi-task_2018, rabinovich_personalized_2017}. Style representations also form a key technique for forensic linguistics, where style representation forms the bedrock for stylometric analysis, enabling authorship verification \cite{boenninghoff_similarity_2019,hay_representation_2020,luar_2021,zhu_idiosyncratic_2021,wegmann_same_2022} and detection of fraudulent texts or malicious activities \cite{manolache_veridark_2022}.

\begin{figure}[t]
  \centering
  \small
  \begin{verbatim}
# 1) Install Biberplus and spaCy
pip install biberplus
python -m spacy download en_core_web_sm

# 2) Minimal usage in Python
from biberplus.tagger import \
    calculate_tag_frequencies

# 3) Tag text and count frequencies
text = "It doesn't seem likely ..."
freq = calculate_tag_frequencies(text)
print(freq)
  \end{verbatim}

  \vspace{1em}
  \includegraphics[width=\linewidth]{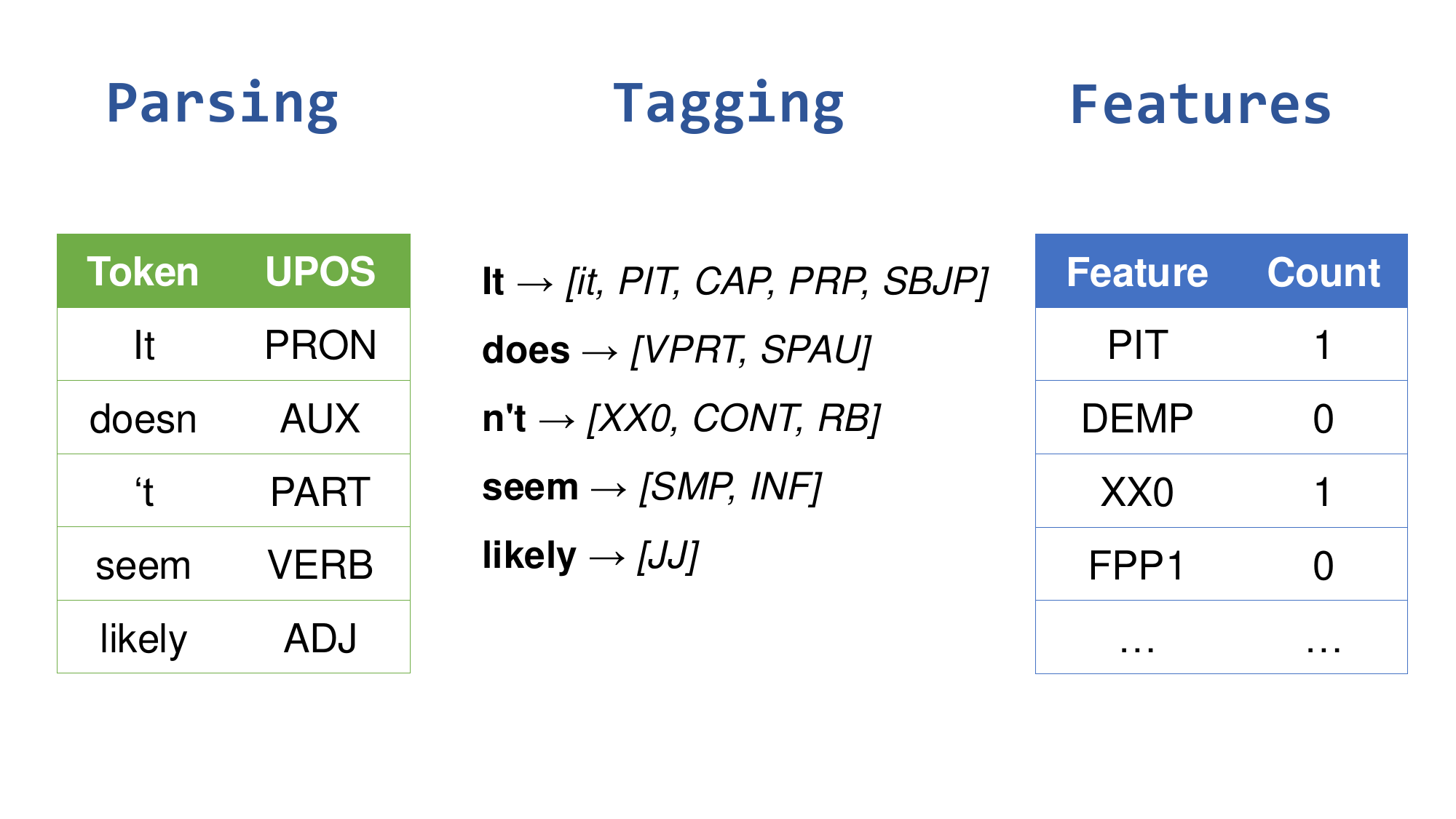}

  \caption{Minimal example of installing and using \biberplus\ }
  \label{fig:code-sample}
\end{figure}

Despite its importance, methods for producing linguistically informed, high-coverage, and scalable style representations remain limited. Existing systems \cite{nini2019multi,wilson10paclic} often rely on small sets of stylistic features or require significant manual effort to scale to large corpora. We address these gaps through two main contributions. First, we introduce \biberplus, a Python toolkit that combines 96 Biber-style features in a single, easy-to-use package, offering both precise, rule-based tagging and integrated analytics (e.g., PCA, factor analysis). Drawing on foundational works in linguistic style profiling \citep{biber_1988,Grieve2018MappingLI,gimpel-etal-2011-part,clarke_dimensions_2017}, \biberplus automates the extraction of syntactic, lexical, and discourse-level style attributes. In short, \biberplus\ aggregates and automates an extensive range of stylistic features to enable straightforward and interpretable style analysis without juggling multiple tools. 
For example, \biberplus can track the frequency of a specific feature (e.g., the \texttt{PIT} tag for Pronoun it) in the snippet shown in \figurename~\ref{fig:code-sample}.

Building upon these resources, we present the \neurobiber\ system---a neural tagger that predicts whether a given \biberplus\ feature is present in a text with macro- and micro-F1 scores of 0.97 and 0.98, respectively. This neural approach achieves speeds at an average rate of 117,000 tokens per second, enabling large-scale tagging scenarios that were previously intractable. To illustrate the impact, consider a massive dataset like Common Crawl containing around 1.1~trillion tokens. Using eight GPUs in parallel, \neurobiber\ would require about 13.6 days to process the entire dataset. In contrast, a CPU-based open-source Multidimensional Analysis Tagger~\cite{nina_2019_MDA} run with the same level of parallelism would take approximately 756 days (over two years). \neurobiber\ addresses a critical gap in the field by combining the interpretability of traditional linguistic feature sets with the scalability and efficiency of neural methods. By enabling rapid, high-fidelity style tagging across massive datasets, it bridges the divide between handcrafted linguistic insights and the data-driven demands of modern NLP. This integration sets the stage for future innovations in explainable, large-scale style analysis across diverse applications.

In the remainder of this paper, we first provide background on linguistic style representations in \Sref{sec:style}, discussing Biber's framework~\citep{biber_1988} and other systems such as LISA \cite{patel_learning_2023} and LUAR \cite{luar_2021}. We describe the design and implementation details of both the Python-based and neural versions of the Biber taggers in \Sref{sec:biberTaggers}. Our experimental setup and evaluation demonstrate the effectiveness and scalability of \neurobiber\ in generating robust, high-fidelity style representations. We conclude by outlining potential future directions and use cases for large-scale stylometric modeling.

\section{Linguistic Style}
\label{sec:style}

Style, broadly, refers to the patterns of lexical, syntactic, and discourse features in text \cite{biber_conrad_2009,crystal_style_1969}. However, defining style remains complex because it can be approached in many ways. In this work, we build on the ideas of \citet{biber_conrad_2019} by using style as a flexible concept that can capture linguistic variation at multiple levels—whether individual (idiolect), group-based, or tied to specific domains. In contrast, register represents the configuration of linguistic forms suited to a particular communicative context, such as news articles or text messages, which exhibit specific levels of formality, conventions, and audiences. While register analysis excels at revealing functional patterns shaped by situational contexts, stylistic analysis highlights how authors or groups may adopt or diverge from those patterns based on various linguistic tendencies. Biber-style features can be leveraged to examine how different authors, groups, or registers choose to realize these forms—making style analysis a complementary approach for both individual and collective language use.

Biber’s Multidimensional Analysis (MDA) \cite{biber_1988,biber_conrad_2009,biber_conrad_2019} has been a highly influential method for examining both style and register. MDA enumerates a broad set of lexical and syntactic features that capture variation such as pronoun use, passive constructions, stance markers, and subordination, combining these features into empirically derived dimensions of linguistic variation. MDA has found wide adoption across diverse text genres \cite{clarke_dimensions_2017,clarke_functional_2019} and inspired multiple open-source taggers and utilities. Three tools—the Multidimensional Analysis Tagger \cite{nini2019multi}, biberpy \cite{sharoff2021genre}, and profiling-ud \cite{brunato2020profiling,de2021universal}—implement Biber’s feature set and demonstrate the usefulness of explicit, interpretable stylistic features for tasks in corpus linguistics and NLP. Although each of these tools offers valuable capabilities, \biberplus\ includes an expanded set of stylistic features, supports GPU-based processing for efficiency, and, through our \neurobiber\ model, can operate up to 56x times faster than comparable libraries.

In parallel with Biber-inspired approaches, neural embedding-based methods aim to learn broad, domain-agnostic representations of style. LUAR \cite{luar_2021} focuses on large-scale authorship verification by combining self-attention with contrastive training, enabling it to handle vast numbers of authors. Meanwhile, LISA \cite{patel_learning_2023} emphasizes interpretability by leveraging GPT-3-generated annotations to create a synthetic stylometric dataset, STYLEGENOME, which trains models to recognize 768 distinct, linguistically interpretable style attributes, such as formality or humor. This allows LISA to capture high-level stylistic trends while providing transparency into the factors shaping its representations. In contrast, Biber-inspired features, such as syntactic constructions (e.g., frequency of passive voice), offer detailed, low-level insights into linguistic style.

\begin{table*}[t]
\resizebox{\textwidth}{!}{
\begin{tabular}{ll cccc}
\textbf{Source}             & \textbf{Description}                                                      & \textbf{Authors} & \textbf{Texts} & \textbf{Avg. Texts per Author} & \textbf{Avg. Text Tokens}\\ 
\hline
Wikipedia editing histories & The edit history from Wikipedia description                               & 506,992          & 3,134,068      & ~~6.18                           & 232 \\ %
Wikipedia discussions       & Article editing discussions from Wikipedia& 188,663          & 4,204,118      & 22.28                          & 116 \\ %
Book3Corpus                 & A large collection of novel books which contains \textbackslash{}bn books & 66,385           & 1,276,515      & 19.22                          & 700                       \\
Amazon reviews              & Amazon product reviews & 128,495          & 1,771,985      & 13.79                          & 143 \\ %
Gmane mailing list          & Gmane is a mailing list archive started in 2002                           & 693,135          & 4,820,910      & ~~6.95                           & 123  \\ %
Realnews article collection & Borpus of news articles from Common Crawl& 212,970          & 15,874,737     & 74.53                          & 609 \\ %
Reddit                      & Reddit comments spanning 15 years& 2,459,016        & 12,290,058     & ~~4.99                           & 117 \\ %
\end{tabular}}
\centering
\caption{Summary of the multi-domain dataset used to train \neurobiber. The dataset includes \~42 million samples across seven  sources, encompassing a wide range of text lengths and styles}
\label{table:datasets}
\end{table*}

\section{Biber Taggers} \label{sec:taggers}\label{sec:biberTaggers}

Extracting linguistic features at scale remains a core challenge when analyzing text style. To address this, we release two complementary tools for Biber-based feature tagging: \biberplus, a rule-based Python implementation that uses explicit patterns, and \neurobiber, a neural approach designed for high-speed, large-scale tagging.

\subsection{BiberPlus Python Implementation}
\label{sec:biberplus}

\biberplus\ advances the original linguistic framework proposed by \citet{biber_1988}, offering an expanded set of features and integrated analytical tools. Implemented entirely in Python, \biberplus\ tracks 96 linguistic features, encompassing Biber’s original 67 tags and additional features derived from recent research \citep{clarke_dimensions_2017,clarke_functional_2019}. A comprehensive list of these features is provided in Appendix~\ref{appendix-biberplus}.  At its core, \biberplus\ first uses SpaCy \citep{honnibal2020spacy} for high-performance parsing and then applies a collection of hand-crafted rules to the resulting syntactic trees. These rules capture the exact constructions outlined by Biber’s framework (e.g., passive voice, subordination, stance markers) and newer expansions, thereby ensuring that each feature count remains faithful to known definitions. We developed 225 unit tests to verify and stress-test these handcrafted patterns in realistic scenarios, guaranteeing robust coverage and accuracy across a range of text types.

\paragraph{Flexibility}
\biberplus\ is designed to accommodate diverse text types, from very short social media posts to lengthy academic articles. While regular frequency counts are appropriate for longer texts, short passages can lead to highly variable or skewed frequencies. Therefore, \biberplus\ also provides a \textbf{binary counting} mode, which captures only the presence or absence of each feature in user-defined chunks (e.g., every 100 tokens). This binary option is especially useful for short texts like social media captions or comments, where raw frequencies are less stable and may not accurately capture stylistic presence. In the regular counting mode, the tool calculates how often each feature appears in user-defined chunks of text—commonly every 1000 tokens—and then aggregates the mean, minimum, maximum, and standard deviation across these chunks. By contrast, the binary counting mode captures only the presence or absence of the feature in each chunk. For instance, if a 500-token text is divided into five 100-token chunks, and a feature appears in two of those chunks, the binary count is $0.4$. Under regular counts, if it appears 10 times total, the average frequency is 2 occurrences per 100 tokens. This distinction allows researchers to fine-tune how stylistic features are counted based on text length and analytic goals. Installation and usage require only a simple \texttt{pip install} command and a few lines of code (see Figure~\ref{fig:code-sample}), and the open-source repository\footnote{\url{https://github.com/davidjurgens/biberplus}} welcomes community contributions. Notably, \biberplus\ is already 2.2x faster than the fastest existing open-source tagger, making it viable even on modest hardware for many text analysis tasks.

\paragraph{Dataset Tagging}
Using the multi-domain dataset summarized in Table~\ref{table:datasets}, we use \biberplus\ to tag all 42 text samples. We ran three \biberplus\ jobs in parallel for two weeks to compute feature frequencies per 100 tokens, producing a comprehensive style profile for each text. To accommodate both short and extended passages, counts and binary indicators were computed simultaneously. 

\subsection{\neurobiber\ Model}
\label{sec:neurobiber}

We develop \neurobiber, a neural tagger that significantly accelerates the extraction of Biber-style features. By leveraging a transformer-based model, \neurobiber\ predicts whether each of the 96 linguistic features tracked by \biberplus\ is present in a given text. This approach achieves speeds up to \textbf{56 times faster} than the fastest open-source system. The model is available as a HuggingFace transformer,\footnote{\url{https://huggingface.co/Blablablab/neurobiber/}} facilitating easy integration into existing NLP pipelines.

\paragraph{Model Architecture and Training}

\neurobiber\ is built upon RoBERTa \citep{liu2019robertarobustlyoptimizedbert}, fine-tuned for multi-label sequence classification. Training data comes from the multi-domain corpus tagged by \biberplus\ (Table~\ref{table:datasets}), with an 80/10/10 split at the chunk level, yielding about 32 million sequences for training, 4 million for development, and 4 million for testing. Each chunk corresponds to 100 tokens, and the label for each of the 96 features is binary (present or absent) based on \biberplus\ tags. In this way, \neurobiber\ directly mimics whether a Biber feature is found in each chunk, retaining interpretability because its predictions align with the same linguistic definitions as \biberplus. For text segments longer than 512 tokens, we process each 512-token span as a separate input; predictions are then aggregated across spans by marking a feature as present if at least one span predicts it. We evaluate performance by comparing predicted labels against the ground-truth labels from \biberplus\ on the held-out test set, measuring both micro- and macro-F1. \neurobiber\ achieves macro-F1 = 0.97 and micro-F1 = 0.98 on this held-out set, demonstrating high fidelity in reproducing \biberplus\ tags. Detailed, per-feature metrics are provided in Appendix~\ref{ref:appendix-neurobiber}.

\neurobiber\ is built upon the RoBERTa \citep{liu2019robertarobustlyoptimizedbert}, fine-tuned for multi-label sequence classification. Training data comes from the multi-domain corpus tagged by \biberplus\ (Table~\ref{table:datasets}), where feature counts were converted into presence/absence labels every 100 tokens with a data split of 80/10/10. In this way, \neurobiber\ directly mimics whether a Biber feature is found in each chunk, retaining interpretability because its predictions align with the same linguistic definitions as \biberplus. For sequences longer than 512 tokens, the text is chunked; predicted features in any chunk are aggregated for the entire text. The model achieves high performance on a held-out test set (macro-F1: 0.97, micro-F1: 0.98). Detailed performance metrics for each feature are provided in Appendix~\ref{ref:appendix-neurobiber}.

\paragraph{Efficiency and Throughput}

\begin{figure}[]
  \centering
  \includegraphics[width=\linewidth]{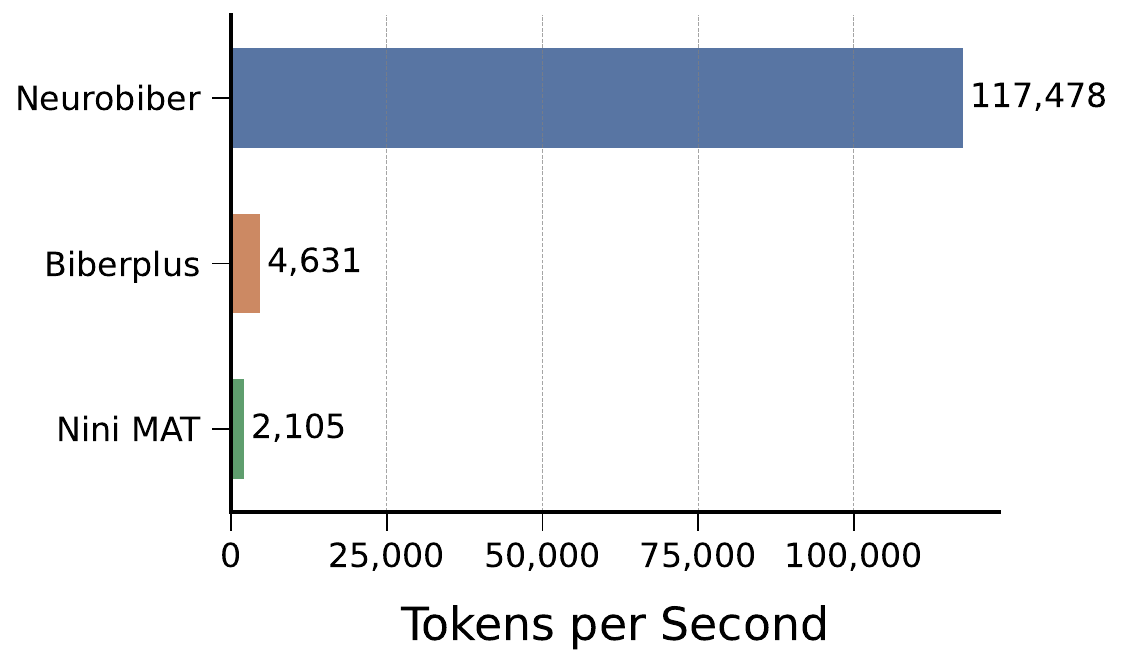}
  \caption{Tagging-speed comparison of three systems on 2.4M tokens. \neurobiber\ achieves the highest throughput, enabling large-scale style analysis at up to 56x the speed of an existing tagger \citep{nina_2019_MDA}}
  \label{fig:biber_performance}
\end{figure}

Figure~\ref{fig:biber_performance} compares the throughput of three systems: an existing open-source multidimensional analysis tagger \citep{nini2019multi} that runs on a single CPU and 67 features, \biberplus, and \neurobiber\ with 96 features, on 7,000 sampled texts (1,000 from each dataset) totaling 2.4 million tokens. \biberplus\ alone is 2.2x faster than the open-source baseline. Meanwhile, \neurobiber\ further boosts tagging speed to about 56x faster than the same baseline, enabling real-time or large-scale batch processing scenarios that were previously infeasible. For \neurobiber, we run on a single NVIDIA RTX A6000 GPU with a batch size of 128.

\paragraph{Recommendation and Applications}

For large-scale datasets, we recommend using \neurobiber\ because its trained model can perform high-fidelity style tagging at massive scales, making it ideal for analyzing large corpora or integrating with downstream machine learning systems. We trained \neurobiber\ specifically on presence/absence signals per 100 tokens, which works well for most short to medium-length texts. However, for tasks requiring exact feature counts, or if one is working with long-form texts that benefit from detailed parsing, \biberplus\ remains a valuable choice, offering precise and customizable feature extraction with integrated analytical capabilities.

\section{Demonstrations}
\label{sec:demonstrations}

In this section, we illustrate how \neurobiber\ combines interpretability with speed for large-scale stylometric analysis. First, we replicate Biber’s original MDA by applying PCA to \neurobiber-extracted features, showing that our neural approach naturally separates text types in ways that closely resemble the classic analysis (\S\ref{sec:biberReplicates}). Then, to demonstrate the practical benefit of these quickly generated features, we show they can be used for an authorship-verification task on the PAN dataset (\S\ref{sec:pan_authorship}), where \neurobiber\ matches parser-based taggers’ performance at a fraction of the time and computational cost.

\begin{figure*}[ht!]
  \centering
  \includegraphics[width=\textwidth]{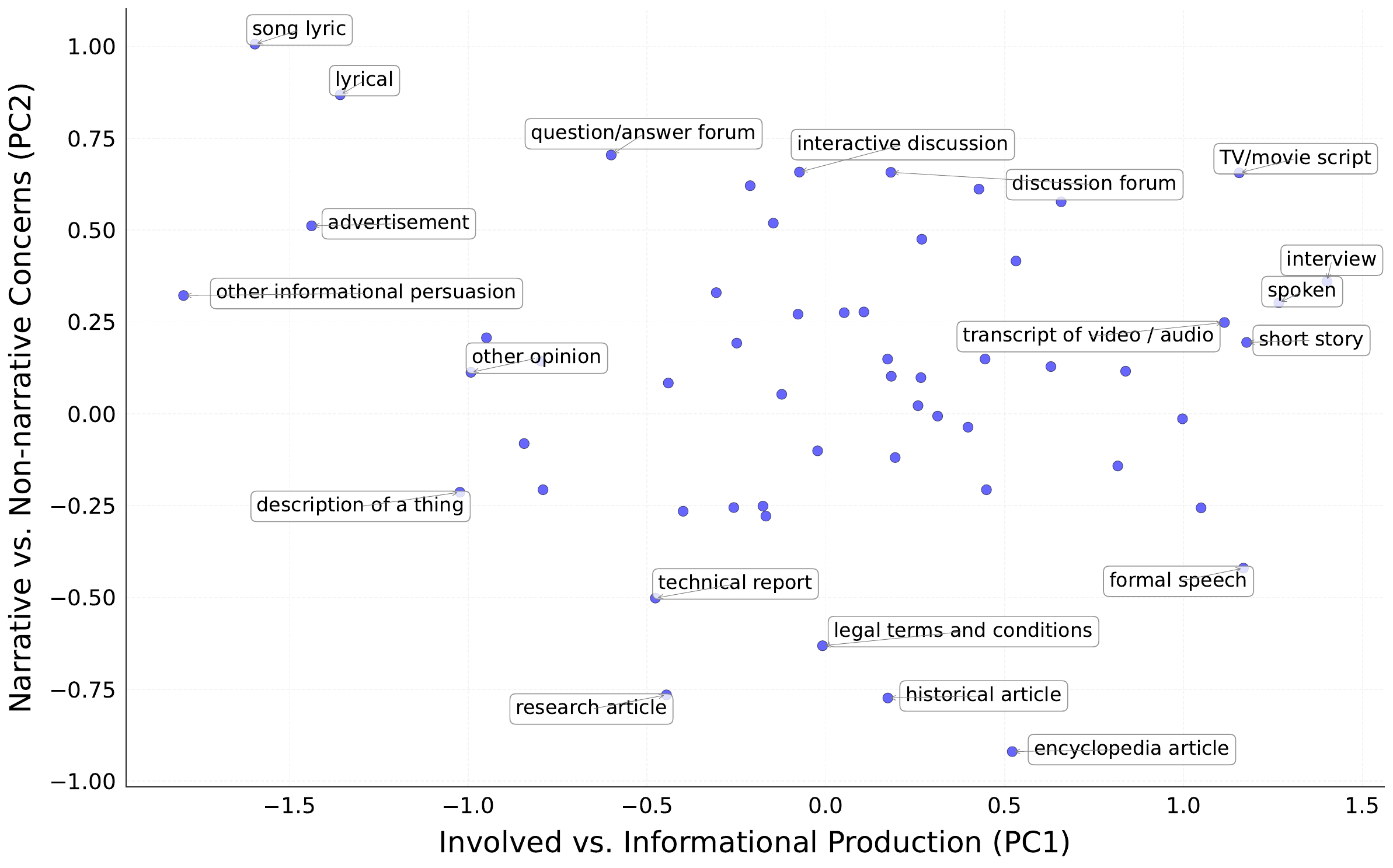}
  \caption{CORE corpus registers plotted along PC1 and PC2 of a PCA on \neurobiber\ features. Clusters reveal an \emph{Involved} vs.\ \emph{Informational} contrast, mirroring Biber’s Dimension~1.}
  \label{fig:register_variation}
\end{figure*}

\subsection{Principal Component Analysis}
\label{sec:biberReplicates}

We apply PCA to the \neurobiber-extracted features across the CORE corpus registers to see if the resulting principal components (PCs) replicate the well-known dimensions from Biber’s MDA framework \citep{biber_1988}. Figure~\ref{fig:register_variation} plots two of these components, illustrating clear, interpretable contrasts among registers. Our main findings include:

\paragraph{PC2 as Involved vs. Informational.}
The second principal component (PC2) primarily distinguishes more involved (often interactive) registers (e.g., discussion forums, Q\&A) from text types that emphasize expository or informational content (e.g., technical reports, encyclopedia articles). In other words, PC2 aligns well with Biber’s Dimension~1, where top-loading features for involved text include personal pronouns and contractions, whereas informational texts rely heavily on nominal and prepositional density. Notably, this dimension can capture either inter-personal style (spoken-like or colloquial) or more objective, formal style.

\paragraph{Other Principal Components}
In some cases, PCA merges stylistic traits that Biber originally treated as separate dimensions. For instance, PC1 groups short stories, speeches, and interviews with certain opinion-oriented texts, suggesting a continuum of narrative or personal stance versus more formal or impersonal presentation. Our inspection of top-weighting features (e.g., first-person pronouns, stance verbs) reveals that narrative and persuasive elements often cluster because both contain direct or subjective expression. Similarly, PC3 and PC4 highlight explicit reference (e.g., nominal references) versus online elaboration (e.g., insertions or parentheticals), occasionally merging text genres (magazine articles, opinion pieces) that share these discourse-level features. Overall, these partial overlaps reflect how factor analyses can conflate multiple stylistic tendencies, especially when the corpus spans written, spoken, and digital registers.

\subsection{Authorship Verification with \neurobiber}
\label{sec:pan_authorship}

To demonstrate the real-world benefits of rapid style tagging, we apply \neurobiber\ to the authorship verification task from PAN 2020 \citep{stein_pan_2020} (small). This task requires determining whether two text samples—often fanfiction that switches style depending on the fictional “universe”—were written by the same individual. Because the texts can vary widely in topic and register, authorship signals must come from stable stylistic patterns rather than content alone. We compare \neurobiber\ to:

\begin{itemize}
    \item \textbf{Random Classifier}: Assigns labels at random.
    \item \textbf{Bi-Encoder Model}: A RoBERTa-based bi-encoder trained on 42,000 text pairs using a contrastive objective. 
\end{itemize}
At inference, the bi-encoder produces two embeddings that we concatenate for classification. By contrast, \neurobiber\ extracts a 96-dimensional feature vector from each text; these vectors are concatenated and fed into a Random Forest. In both cases, the final classification model is trained on the PAN 2020 data.

\paragraph{Results and Analysis}
\begin{table}[t]
    \centering
    \begin{tabular}{l c}
    \textbf{Model} & \textbf{F1 Score} \\
    \hline
    Random Classifier & 0.50 \\
    RoBERTA Bi-Encoder & 0.78 \\
    \neurobiber\ + Random Forest & 0.77 \\
    \end{tabular}
    \caption{Performance on the PAN 2020 authorship verification task (small). \neurobiber\ achieves an F1 score comparable to the fine-tuned bi-encoder model, significantly outperforming the random baseline.}
    \label{tab:model_performance}
\end{table}

Table~\ref{tab:model_performance} shows each model’s F1 score. \neurobiber\ achieves 0.77, nearly matching the fine-tuned bi-encoder (0.78) and far exceeding the random baseline (0.50). Notably, generating the \neurobiber\ feature vectors for the entire dataset took only about half an hour—much less than the time and GPU resources typically required to train or re-train large transformer models. These concise style vectors may complement deep learning representations by capturing stable, interpretable signals (e.g., personal pronoun frequencies, modal usage) that large neural models may not explicitly expose. In settings where interpretability or resource constraints are paramount, \neurobiber\ offers a practical alternative or addition to deep embeddings, demonstrating that well-crafted stylistic features can stand alongside more complex representations in tasks like authorship verification.

\subsection{Future Applications}

\neurobiber’s performance shows that carefully designed stylistic features can function as a stable “fingerprint” for authorship, even when texts differ significantly in topic or register. Unlike purely content-based cues, these linguistically grounded signals remain consistent across domains, providing deeper syntactic or lexical insights into an author’s style. This consistency is especially valuable in forensic or literary contexts, where understanding why two texts are stylistically similar matters as much as whether they match. 
Beyond accuracy, \neurobiber’s reduced reliance on parser-based pipelines lowers the computational overhead for large-scale style monitoring. Researchers can thus analyze massive corpora—from historical archives to social media streams—at speeds that enable repeated or near real-time updates. This capability opens new avenues for tracking stylistic shifts across genres or time periods and for practical applications like social media trend analysis, where classical parsing methods would be prohibitively slow.
Additionally, \neurobiber’s interpretable feature vectors can be combined with deep neural embeddings, merging granular stylistic cues with semantic information. This fusion facilitates personalization (e.g., recommendation systems) and domain adaptation when fine-tuning large language models on specialized corpora. Instead of parsing every text, \neurobiber\ can efficiently scan millions of samples and highlight those that align stylistically with a target domain, making downstream training more focused and transparent.

\section{Conclusion}
In this paper, we introduced \neurobiber, a neural approach for fast and interpretable stylistic feature extraction based on Biber’s Multidimensional Analysis framework. By leveraging transformer-based models, \neurobiber\ predicts the presence of 96 linguistic features with high accuracy—achieving speeds up to 56 times faster than parser-based systems. We also presented \biberplus, an enhanced Python implementation offering flexible counting methods and integrated analytical tools such as PCA and factor analysis. Experimental results showed that \neurobiber\ effectively replicates key findings from traditional linguistic analyses, recovering Biber-style dimensions of variation. Additionally, on an authorship verification task, \neurobiber\ delivered competitive performance without extensive retraining, underscoring its practical value for real-world applications.

\clearpage

\section*{Ethical Considerations}

When applied to authorship verification and related stylometric tasks, \neurobiber\ raises important privacy considerations. Although our methods can aid forensic or security investigations, they may also be used maliciously to identify or profile individuals without their consent. By revealing distinct linguistic patterns, \neurobiber\ could inadvertently compromise a writer’s anonymity, particularly in sensitive domains such as whistleblowing or personal communications. Consequently, deploying these techniques requires careful adherence to legal and ethical guidelines, including transparent data policies and clear user consent.  

Our research contributes to responsible use by enhancing model interpretability, allowing practitioners to scrutinize which features drive decisions and thus detect potential bias or misuse. Future work might build on these interpretability insights to develop adversarial defenses that protect user identities. All data used in our experiments were publicly available and processed under applicable terms of service; authors were anonymized to the extent possible. We encourage continued dialogue among technologists, policymakers, and stakeholders to ensure that stylometric technologies like \neurobiber\ serve legitimate, ethical purposes and respect individuals’ right to privacy.

\bibliography{references,anthology}
\bibliographystyle{acl_natbib}
\clearpage
\appendix
\onecolumn

\section{Biberplus}\label{appendix-biberplus}

\small
\label{tab:biberplus-features}
\begin{supertabular}{|p{1.1cm}|p{3.2cm}|p{1.1cm}|p{3.2cm}|p{1.1cm}|p{3.2cm}|}
\hline
\textbf{Feat.} & \textbf{Description} &
\textbf{Feat.} & \textbf{Description} &
\textbf{Feat.} & \textbf{Description}\\ 
\hline
VBD & Past tense POS 
 & PEAS & Perfect aspect using patterns like HAVE + (ADV) + (ADV) + VBD/VBN 
 & VPRT & Present tense: VBP or VBZ tag\\
\hline
PLACE & Any item in the place adverbials list that is not a proper noun (NNP) 
 & TIME & Time adverbials excluding `soon` when followed by `as`
 & FPP1 & First person pronouns\\
\hline
SPP2 & Second person pronouns 
 & TPP3 & Third person pronouns 
 & PIT & Pronoun `it` and its related forms\\
\hline
INPR & Indefinite pronouns list 
 & DEMP & Demonstrative pronouns under certain conditions
 & PROD & Pro-verb `do` excluding specific patterns\\
\hline
WHQU & Direct WH-questions 
 & NOMZ & Noun endings like -tion, -ment, -ness
 & GER & Gerunds of length greater than or equal to 10 ending in ing or ings\\
\hline
NN & Total other nouns (not nominalisations or gerunds) 
 & PASS & Agentless passives 
 & BYPA & By-passives following `PASS` tag\\
\hline
BEMA & Be as the main verb 
 & EX & Existential `there` 
 & THVC & That-verb complements with specific preceding or succeeding patterns\\
\hline
PASTP & Past participial clauses: punctuation + VBN
 & WZPAST & Past participial WHIZ-deletion relatives 
 & WZPRES & Present participial WHIZ-deletion relatives: VBG preceded by an NN\\
\hline
TSUB & That-rel. clauses on subject position
 & TOBJ & That-rel. clauses on object position
 & WHSUB & WH-rel. clauses on subject position\\
\hline
WHOBJ & WH rel. clauses on object position
 & PIRE & Pied-piping relatives (prep + whom/who/whose/which)
 & SERE & Sentence relatives: punctuation + `which`\\
\hline
CAUS & Any occurrence of the word "because"
 & CONC & Any occurrence of "although", "though", "tho"
 & COND & Any occurrence of "if" or "unless"\\
\hline
OSUB & Other adverbial subordinators (multi-word)
 & PIN & Total prepositional phrases
 & JJ & Attributive adjectives\\
\hline
PRED & Predicative adjectives
 & RB & Adverbs (POS tags RB, RBS, RBR, WRB)
 & CONJ & Conjuncts with preceding punctuation or specific words\\
\hline
DWNT & Specific downtoners
 & AMP & Specific amplifiers
 & DPAR & Discourse particle: pattern preceded by punctuation\\
\hline
HDG & Hedge words
 & EMPH & Emphatics
 & DEMO & Demonstratives\\
\hline
POMD & Possibility modals
 & NEMD & Necessity modals
 & PRMD & Predictive modals\\
\hline
PUBV & Public verbs
 & PRIV & Private verbs
 & SUAV & Suasive verbs\\
\hline
SMP & Seem/Appear verbs
 & CONT & Reduced forms (contractions)
 & THATD & Subordinator “that” deletion\\
\hline
STPR & Stranded preposition
 & SPIN & Split infinitives
 & SPAU & Split auxiliaries\\
\hline
PHC & Phrasal coordination
 & ANDC & `and` in coordination of independent clauses
 & XX0 & Words indicating analytic negation (`not`)\\
\hline
SYNE & Synthetic negations before adj./noun
 & QUAN & Quantifier words
 & QUPR & Quantifier pronouns\\
\hline
ART & Articles (`the`, `a`, `an`)
 & AUXB & Auxiliary verbs that are forms of `be`
 & CAP & Words starting with a capital letter\\
\hline
SCONJ & Subordinating conjunctions
 & CCONJ & Coordinating conjunctions
 & DET & Determiners\\
\hline
EMOJ & Emojis
 & EMOT & Common emoticons
 & EXCL & Exclamation mark `!`\\
\hline
HASH & Words starting with \#
 & INF & Infinitive verb forms
 & UH & Interjections\\
\hline
NUM & Numerals
 & LAUGH & Laughter acronyms like `lol`
 & PRP & Possessive pronouns\\
\hline
PREP & Prepositions
 & NNP & Proper nouns
 & QUES & Question mark `?`\\
\hline
QUOT & Quotation marks
 & AT & Words starting with `@`
 & SBJP & Subject pronouns\\
\hline
URL & URLs
 & WH & WH words (`who`, `what`, `where`)
 & INDA & Indefinite articles (`a`, `an`)\\
\hline
ACCU & Accusative case pronouns
 & PGAS & Progressive aspect verb forms
 & CMADJ & Comparative adjectives\\
\hline
SPADJ & Superlative adjectives
 & X & Words not fitting other POS categories
 & AWL & Mean word length\\
\hline
TTR & Type-token ratio
 & TRB & Total adverbs
 & & \\
\hline
\end{supertabular}

\clearpage

\section{Neurobiber} \label{ref:appendix-neurobiber}

\subsection{Training Details}
We trained a RoBERTa-base model in a multi-label classification setup (96 labels) using \texttt{adamw\_torch} with a learning rate of \texttt{2e-5}, \(\beta_1=0.9\), \(\beta_2=0.999\), an \(\epsilon\) of \texttt{1e-8}, and a weight decay of \texttt{0.01}. Each mini-batch contained 256 sequences, and training was performed for 8~epochs on a dataset of approximately 31.7~million examples, with a dev and test size of about 4~million examples each. We use gradient checkpointing to reduce memory usage and used bf16 precision throughout training; the random seed was fixed at 42 to ensure reproducibility. A linear scheduler with no warmup steps was applied, and the best checkpoint was selected via validation.

\clearpage
\subsection{Performance by Tag}
\label{tab:f1-support}
\begin{supertabular}{|p{1.1cm}|p{5.5cm}|p{1.0cm}|p{2.0cm}|}
\hline
\textbf{Acronym} & \textbf{Full Name} & \textbf{F1} & \textbf{Train Support} \\
\hline
NN & Total other nouns not tagged as a nominalisation or a gerund & 0.999 & 3.96e+6 \\
\hline
CAP & Tags words starting with a capital letter & 0.999 & 3.95e+6 \\
\hline
PREP & Identifies prepositions & 0.999 & 3.95e+6 \\
\hline
PIN & Total prepositional phrases & 0.999 & 3.95e+6 \\
\hline
ART & Identifies articles like 'the', 'a', 'an' & 0.999 & 3.93e+6 \\
\hline
DET & Identifies determiners & 0.999 & 3.95e+6 \\
\hline
AUXB & Tags auxiliary verbs that are forms of 'be' & 0.999 & 3.82e+6 \\
\hline
CONJ & Conjuncts with preceding punctuation or specific words & 0.999 & 3.50e+6 \\
\hline
INDA & Identifies indefinite articles ('a', 'an') & 0.999 & 3.64e+6 \\
\hline
PRP & Tags possessive pronouns & 0.999 & 3.87e+6 \\
\hline
SBJP & Labels subject pronouns & 0.999 & 3.87e+6 \\
\hline
CCONJ & Labels coordinating conjunctions & 0.999 & 3.81e+6 \\
\hline
JJ & Attributive adjectives & 0.998 & 3.93e+6 \\
\hline
PRIV & Private Verbs & 0.998 & 3.29e+6 \\
\hline
VPRT & Present tense: VBP or VBZ tag & 0.997 & 3.87e+6 \\
\hline
TPP3 & Third person pronouns & 0.997 & 2.63e+6 \\
\hline
RB & Any adverb i.e. POS tags RB, RBS, RBR, WRB & 0.996 & 3.83e+6 \\
\hline
PIT & Pronoun 'it' and its related forms & 0.996 & 3.08e+6 \\
\hline
CONT & Reduced Forms (Contractions) & 0.995 & 2.55e+6 \\
\hline
INF & Tags infinitive verb forms & 0.995 & 3.76e+6 \\
\hline
TO & No description available & 0.995 & 3.30e+6 \\
\hline
PUBV & Public Verbs & 0.995 & 2.47e+6 \\
\hline
WH & Tags WH words like 'who', 'what', 'where' & 0.995 & 3.25e+6 \\
\hline
QUAN & Labels quantifier words & 0.995 & 2.87e+6 \\
\hline
SCONJ & Recognizes subordinating conjunctions & 0.995 & 3.42e+6 \\
\hline
NOMZ & Noun endings like -tion, -ment, -ness & 0.995 & 2.97e+6 \\
\hline
BEMA & Be as the main verb & 0.994 & 3.42e+6 \\
\hline
PGAS & Recognizes progressive aspect verb forms & 0.993 & 3.39e+6 \\
\hline
FPP1 & First person pronouns & 0.993 & 3.18e+6 \\
\hline
PEAS & Perfect aspect using patterns like HAVE + (ADV) + (ADV) + VBD/VBN & 0.992 & 3.40e+6 \\
\hline
XX0 & Tags words indicating analytic negation like 'not' & 0.990 & 2.66e+6 \\
\hline
DEMO & Demonstratives & 0.990 & 2.73e+6 \\
\hline
TIME & Time adverbials excluding 'soon' when followed by 'as' & 0.988 & 1.95e+6 \\
\hline
POMD & Possibility Modals & 0.988 & 2.20e+6 \\
\hline
PRMD & Predictive Modals & 0.988 & 2.24e+6 \\
\hline
ACCU & Tags accusative case pronouns & 0.987 & 2.11e+6 \\
\hline
EMPH & Identifies emphatics words & 0.987 & 2.63e+6 \\
\hline
SPP2 & Second person pronouns & 0.987 & 1.92e+6 \\
\hline
NUM & Tags numeral words & 0.987 & 2.95e+6 \\
\hline
QUOT & Identifies quotation marks & 0.987 & 1.49e+6 \\
\hline
PASS & Agentless passives & 0.986 & 2.43e+6 \\
\hline
SUAV & Suasive Verbs & 0.985 & 1.88e+6 \\
\hline
VBD & Past tense POS & 0.985 & 3.15e+6 \\
\hline
NNP & Tags proper nouns & 0.982 & 3.56e+6 \\
\hline
X & Tags words not fitting other parts of speech categories & 0.981 & 2.61e+6 \\
\hline
COND & Any occurrence of the words "if" or "unless" & 0.981 & 1.65e+6 \\
\hline
QUES & Labels the question mark '?' & 0.980 & 1.24e+6 \\
\hline
INPR & Indefinite pronouns list & 0.980 & 1.53e+6 \\
\hline
ANDC & Labels 'and' in coordination within independent clauses & 0.980 & 2.24e+6 \\
\hline
QUPR & Tags quantifier pronouns & 0.978 & 1.37e+6 \\
\hline
PROD & Pro-verb 'do' excluding specific patterns & 0.978 & 2.20e+6 \\
\hline
DEMP & Demonstrative pronouns under certain conditions & 0.978 & 2.12e+6 \\
\hline
THATD & Subordinator "that" Deletion & 0.977 & 1.80e+6 \\
\hline
LAUGH & Recognizes laughter-related acronyms like 'lol' & 0.975 & 4.67e+4 \\
\hline
CMADJ & Tags comparative adjectives & 0.975 & 2.01e+6 \\
\hline
OSUB & Other adverbial subordinators with specific multi-word units & 0.975 & 1.19e+6 \\
\hline
DWNT & Any instance of specific downtoners & 0.975 & 1.28e+6 \\
\hline
THVC & That verb complements with specific preceding or succeeding patterns & 0.974 & 1.32e+6 \\
\hline
EXCL & Tags the exclamation mark '!' & 0.974 & 6.04e+5 \\
\hline
SPADJ & Identifies superlative adjectives & 0.972 & 1.49e+6 \\
\hline
PRED & Predicative adjectives with specific conditions & 0.972 & 2.08e+6 \\
\hline
SPAU & Split Auxiliaries & 0.972 & 1.88e+6 \\
\hline
PLACE & Any item in the place adverbials list that is not a proper noun (NNP) & 0.971 & 1.61e+6 \\
\hline
AMP & Any instance of specific amplifiers & 0.970 & 1.08e+6 \\
\hline
CAUS & Any occurrence of the word "because" & 0.970 & 8.59e+5 \\
\hline
NEMD & Necessity Modals & 0.969 & 8.60e+5 \\
\hline
SMP & Seem/Appear Verbs & 0.967 & 6.62e+5 \\
\hline
STPR & Stranded Preposition & 0.966 & 1.45e+6 \\
\hline
EX & Existential 'there' & 0.965 & 1.23e+6 \\
\hline
PHC & Phrasal Coordination & 0.963 & 2.09e+6 \\
\hline
SERE & Sentence relatives: A punctuation mark followed by the word "which" & 0.963 & 8.60e+5 \\
\hline
WHSUB & WH relative clauses on subject position with specific structure & 0.963 & 1.68e+6 \\
\hline
CONC & Any occurrence of the words "although", "though", "tho" & 0.959 & 5.72e+5 \\
\hline
TSUB & That relative clauses on subject position & 0.959 & 1.20e+6 \\
\hline
BYPA & By-passives following 'PASS' tag & 0.956 & 7.29e+5 \\
\hline
WHCL & No description available & 0.956 & 7.01e+5 \\
\hline
WHQU & Direct WH-questions & 0.955 & 1.06e+6 \\
\hline
SYNE & Identifies synthetic negations followed by adjective, noun, or proper noun & 0.953 & 9.06e+5 \\
\hline
URL & Tags URLs in the text & 0.950 & 2.90e+5 \\
\hline
PIRE & Pied-piping relatives clauses: Preposition followed by whom, who, whose, or which & 0.949 & 3.44e+5 \\
\hline
TOBJ & That relative clauses on object position & 0.945 & 8.12e+5 \\
\hline
HASH & Labels words starting with the \# symbol & 0.943 & 1.09e+5 \\
\hline
HDG & Captures hedge words & 0.942 & 4.42e+5 \\
\hline
WZPRES & Present participial WHIZ deletion relatives: VBG preceded by an NN & 0.941 & 1.54e+6 \\
\hline
SPIN & Split Infinitives & 0.930 & 2.38e+5 \\
\hline
WZPAST & Past participial WHIZ deletion relatives with specific structure & 0.927 & 1.03e+6 \\
\hline
EMOT & Labels common emoticons & 0.925 & 9.21e+4 \\
\hline
PRESP & No description available & 0.922 & 1.08e+6 \\
\hline
THAC & No description available & 0.920 & 3.42e+5 \\
\hline
WHOBJ & WH relative clauses on object position with specific structure & 0.918 & 1.09e+6 \\
\hline
AT & Tags words starting with the '@' symbol & 0.916 & 9.81e+4 \\
\hline
UH & Identifies interjections & 0.913 & 1.04e+6 \\
\hline
GER & Gerunds with length greater or equal to 10 ending in ing or ings & 0.910 & 7.45e+5 \\
\hline
PASTP & Past participial clauses: punctuation followed by VBN & 0.878 & 5.31e+5 \\
\hline
EMOJ & Tags emojis in the text & 0.842 & 1.08e+4 \\
\hline
DPAR & Discourse particle: Specific words preceded by a punctuation mark & 0.832 & 2.82e+5 \\
\hline
\end{supertabular}

\clearpage

\subsection{Register Variation} 
\label{appendix-register-variation}

\begin{figure*}[!htb]
  \centering
  \includegraphics[width=0.6\textwidth]{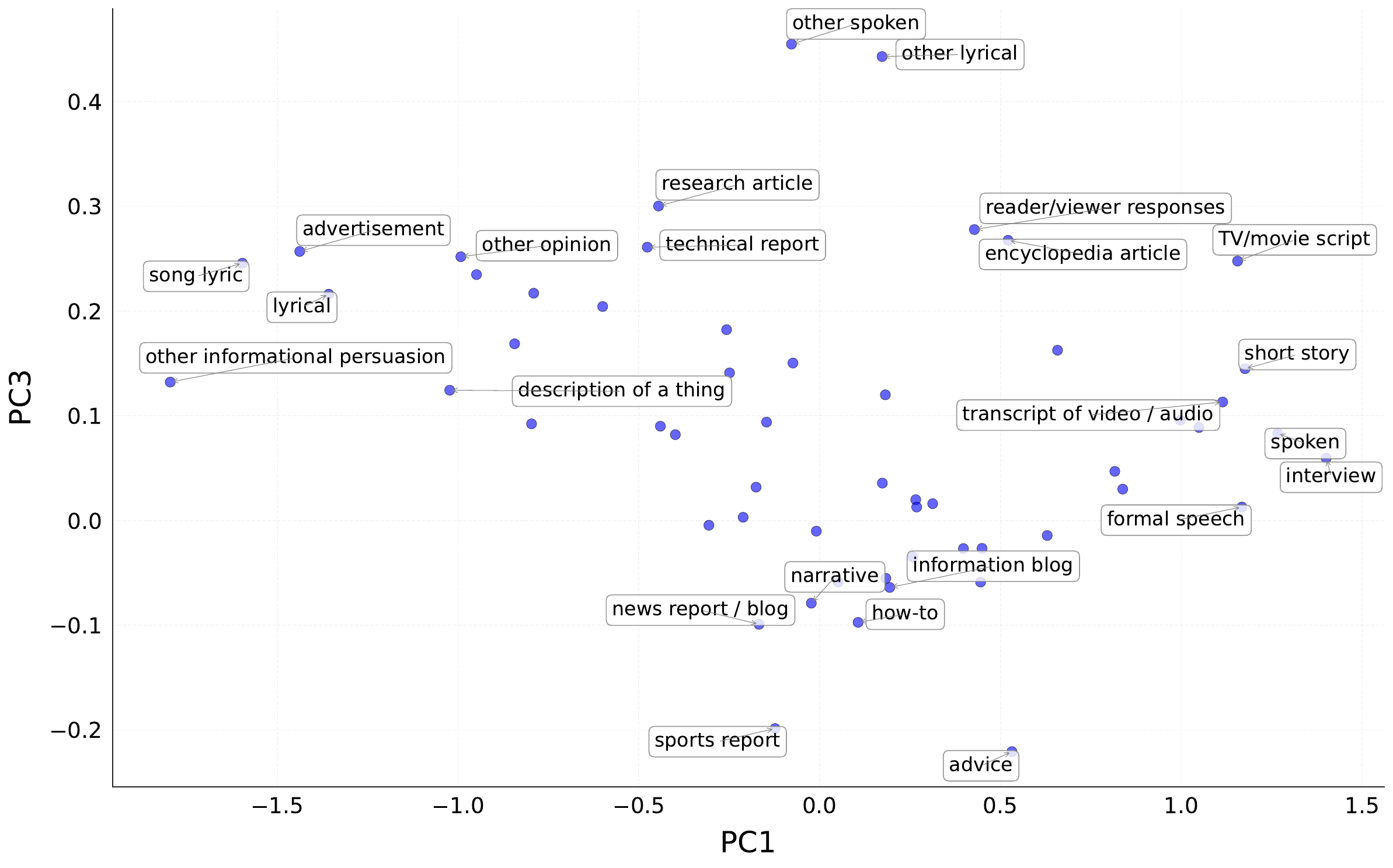}
  \caption{Distribution of registers along Principal Components \textbf{PC1} and \textbf{PC3}.}
  \label{fig:register_variation_pc1_pc3}
\end{figure*}

\begin{figure*}[!htb]
  \centering
  \includegraphics[width=0.6\textwidth]{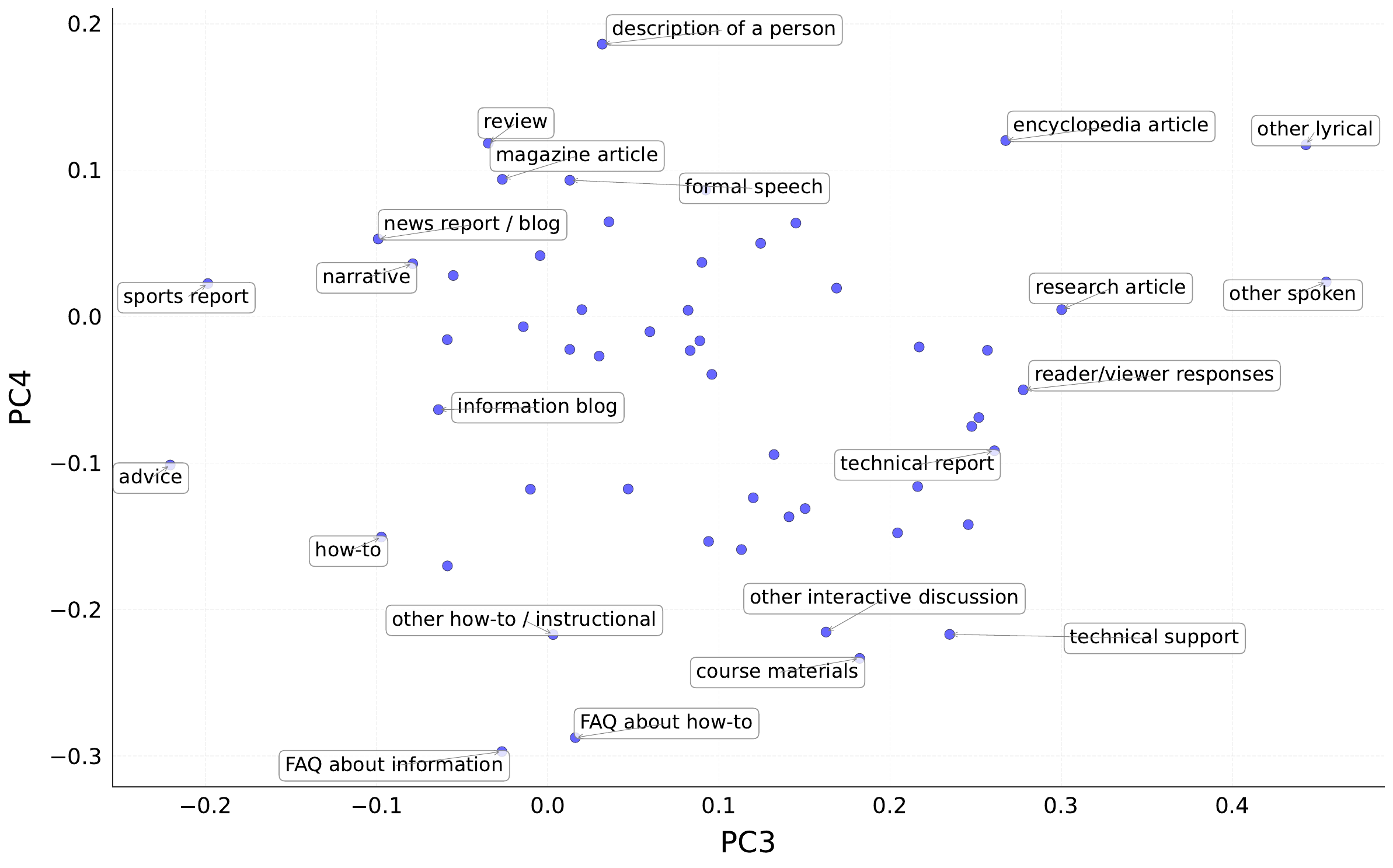}
  \caption{Distribution of registers along Principal Components \textbf{PC3} and \textbf{PC4}.}
  \label{fig:register_variation_pc3_pc4}
\end{figure*}

\begin{figure*}[!htb]
  \centering
  \includegraphics[width=0.6\textwidth]{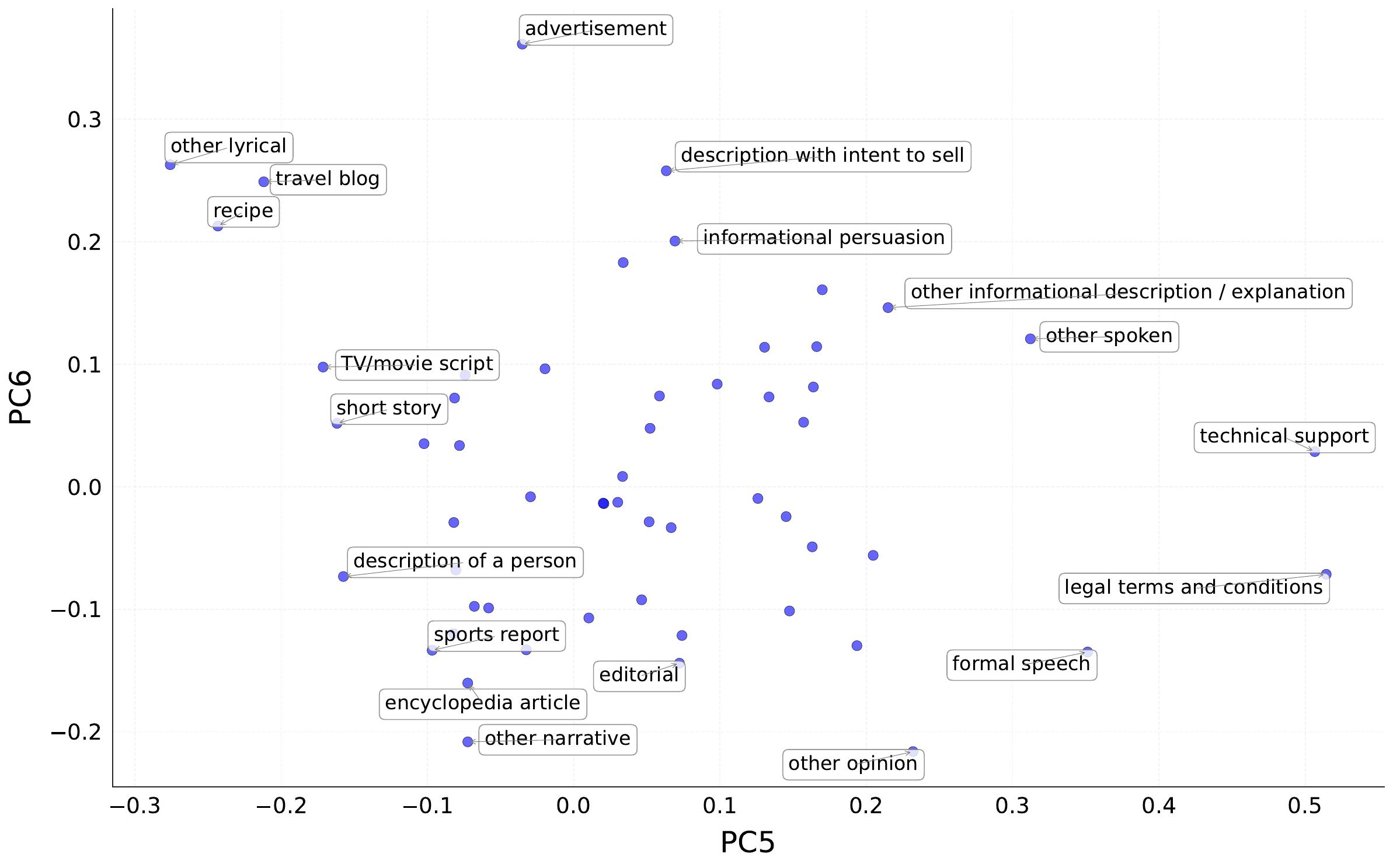}
  \caption{Distribution of registers along Principal Components \textbf{PC5} and \textbf{PC6}.}
  \label{fig:register_variation_pc5_pc6}
\end{figure*}

Figure \ref{fig:register_variation_pc1_pc3} through 
Figure \ref{fig:register_variation_pc5_pc6} illustrate 
the distribution of registers from the CORE corpus 
using \textit{NeuroBiber}. Each subfigure plots two 
principal components, revealing clusters that align 
with Biber's dimensions. For instance, PC1 shows an 
\emph{Involved vs. Informational Production} tendency 
in the data, echoing Biber's original analyses.

\end{document}